\documentclass[letterpaper, 10 pt, conference]{ieeeconf}  
\usepackage{amssymb}
\usepackage{amsmath}
\usepackage{booktabs} 
\usepackage{gensymb}
\usepackage{stfloats}
\usepackage{multirow}
\usepackage{color} 
\usepackage{textcomp}

\IEEEoverridecommandlockouts                              

\usepackage{graphics} 
\usepackage{epsfig} 
\usepackage{mathptmx} 
\usepackage{times} 
\usepackage{hyperref}
\usepackage{cite}
\usepackage{bm}
\title{\LARGE \bf
SGLoc: Semantic Localization System for Camera Pose Estimation from 3D Gaussian Splatting Representation
}

\author{Beining Xu\textsuperscript{\dag1}, Siting Zhu\textsuperscript{\dag1}, Hesheng Wang\textsuperscript{1}
\thanks{\textsuperscript{\dag}The first two authors contributed equally.}
\thanks{*This work was supported in part by the Natural Science Foundation of China under Grant 62225309, U24A20278, 62361166632, U21A20480 and 62403311. Corresponding Author: Hesheng Wang ({\tt\small wanghesheng@sjtu.edu.cn}).}
\thanks{$^{1}$School of Automation and Intelligent Sensing, Key Laboratory of System Control and Information Processing, Ministry of Education of China, Shanghai Jiao Tong University, Shanghai.}%
}

\begin{document}

\maketitle
\thispagestyle{empty}
\pagestyle{empty}

\begin{abstract}

We propose SGLoc, a novel localization system that directly regresses camera poses from 3D Gaussian Splatting (3DGS) representation by leveraging semantic information. 
Our method utilizes the semantic relationship between 2D image and 3D scene representation to estimate the 6DoF pose without prior pose information.
In this system, we introduce a multi-level pose regression strategy that progressively estimates and refines the pose of query image from the global 3DGS map, without requiring initial pose priors.
Moreover, we introduce a semantic-based global retrieval algorithm that establishes correspondences between 2D (image) and 3D (3DGS map).
By matching the extracted scene semantic descriptors of 2D query image and 3DGS semantic representation, we align the image with the local region of the global 3DGS map, thereby obtaining a coarse pose estimation.
Subsequently, we refine the coarse pose by iteratively optimizing the difference between the query image and the rendered image from 3DGS.
Our SGLoc demonstrates superior performance over baselines on 12scenes and 7scenes datasets, showing excellent capabilities in global localization without initial pose prior. Code will be available at \href{https://github.com/IRMVLab/SGLoc}{https://github.com/IRMVLab/SGLoc}.


\end{abstract}

\section{INTRODUCTION}

Visual localization is a fundamental challenge in autonomous driving~\cite{liu2024dvlo, sha2024towards} and robotics~\cite{zhu2024sni}. It enables estimation of 6DoF camera poses within a previously mapped environment. 
Existing traditional localization systems can be categorized into feature-based and regression-based methods. Feature-based methods typically extract 2D and 3D keypoints, then match 2D keypoints from query images with 3D keypoints of the scene~\cite{dusmanu2019d2,lindenberger2023lightglue,sarlin2019coarse} to regress the camera pose. 
Regression-based methods employ neural networks to extract image features and encode absolute poses or scene coordinates for direct 6DoF pose regression~\cite{kendall2015posenet,dong2022visual,li2020hierarchical}.
These methods rely on low-level visual features, such as textural and geometric features. However, low-level visual features are inherently sensitive to environmental variations, particularly in scenes with insufficient texture information or varying lighting conditions, which leads to decreased localization accuracy.



3D Gaussian Splatting (3DGS)~\cite{kerbl20233d} emerges as a promising scene representation. As 3DGS has demonstrated its effectiveness in scene modeling for robotics tasks~\cite{zhu20243d,zhu2024semgauss}, enabling direct pose estimation from 3DGS maps becomes crucial. 
Existing works leverage the high-quality novel view synthesis capability of 3DGS representation to achieve visual localization from 3DGS maps. 
Among these approaches, \cite{sun2023icomma} leverages the rendering process of 3DGS for pose estimation. However, these methods struggle when given poor pose priors, such as significant rotations and translations, leading to substantial discrepancies between the rendering and query views. Such discrepancies result in degraded accuracy of the pose regression.
\cite{liu2024gsloc,zhai2024splatloc,sidorov2024gsplatloc} directly follow the approach of traditional feature-based localization, where keypoints are extracted and matched between 3DGS maps and query images to regress poses.
Consequently, these methods inherit the limitations of traditional localization approaches discussed above.
Furthermore, existing methods overlook the consistency of semantic information between 2D query image and 3D scene representation, resulting in degraded localization performance in complex scenes.

To address these challenges, we propose a novel semantic-based visual localization framework. Our method introduces a multi-level pose regression strategy that integrates semantic-based global retrieval with rendering-based optimization, which enables precise localization of a query RGB image without requiring initial poses. We utilize semantic information to compensate for the inherent shortcomings of traditional feature-based methods.
Specifically, we leverage semantic consistency 
to directly retrieve the closest match to query image from 3DGS map, thereby obtaining a coarse pose estimation. This strategy enables more reliable initial pose estimates for further pose refinement, even in scenes with sparse textural features.
Subsequently, we iteratively optimize the initial pose by comparing rendered images and query images, achieving accurate localization results.

Overall, we provide the following contributions:
\begin{itemize}
\item We present SGLoc, a novel semantic-based localization system that directly regresses camera poses from 3D Gaussian Splatting. 
We introduce a multi-level pose regression strategy that progressively estimates and refines the pose of 2D query image based on the global 3DGS map, without requiring initial pose priors.

\item We employ a semantic-based global retrieval method to establish correspondence between 2D query image and 3DGS semantic representation, thereby obtaining pose estimation of image.

\item Extensive evaluations are conducted on 12Scenes and 7Scenes datasets, to demonstrate the effectiveness of our method in localization performance.

\end{itemize}

\section{Related Work}


\subsection{Traditional Localization} 
Classical localization methods include feature-based methods and regression-based methods. Feature-based methods typically focus on matching keypoints from 2D images and 3D models, then apply Perspective-n-Point (PnP)~\cite{gao2003complete} algorithm with RANSAC~\cite{fischler1981random} for pose estimation~\cite{sarlin2020superglue,sun2021loftr,potje2024xfeat,brachmann2021visual,zhou2022hfnet}. But these methods are easily affected by noise. Regression-based methods employ neural networks to extract image features and encode camera poses or scene coordinates for 6DoF pose regression~\cite{sarlin2021back,kendall2015posenet,chen2022dfnet,kendall2017geometric,shavit2021learning}. Although regression-based methods are faster, they are not superior in accuracy and generalization. Nevertheless, their lack of geometric constraints leads to lower accuracy compared to feature-based methods.

However, due to the geometric ambiguities between 2D and 3D representation, few studies have focused on direct 2D-3D matching.
Given the semantic consistency between 3D and 2D representations, we propose a semantic-based 2D image-to-3DGS map matching method. By aligning the semantic features of query images with known 3DGS map, this method provides reliable initial pose estimation. Our method fully exploits semantic consistency between images and 3D scenes, which significantly enhances the robustness of localization in complex scenarios.

\subsection{NeRF-based Localization}
Neural Radiance Fields~\cite{mildenhall2021nerf} has been utilized for localization tasks for its ability to synthesize novel view images.
iNerf~\cite{yen2021inerf} introduces an inverse NeRF method to estimate camera poses. NeFeS~\cite{chen2023neural} optimizes differences between rendered images and query images to obtain poses. Most approaches~\cite{zhao2024pnerfloc,moreau2023crossfire,zhou2024nerfect,chen2024pgsr} follow traditional feature-based localization methods to match 2D and 3D features. PNeRFLoc~\cite{zhao2024pnerfloc} introduces warping loss to improve pose estimation. NeRFMatch~\cite{zhou2024nerfect} achieves 2D-3D matches with specialized feature extractors. However, those NeRF-based methods all suffer from poor rendering quality and extensive rendering time.

\subsection{3DGS-based Localization}
3D Gaussian Splatting~\cite{kerbl20233d} achieves high quality and real-time novel-view synthesis of the 3D scenes and has recently been employed for visual localization tasks.
Some approaches design the pose estimation framework by combining the rendering process of 3DGS. iComMa~\cite{sun2023icomma} designs a gradient-based differentiable framework to adopt iterative optimization for camera pose regression. 6DGS~\cite{matteo20246dgs} avoids the iterative process by inverting the 3DGS rendering process for direct 6-DoF pose estimation. However, both of them struggle when given poor initial poses, like large rotations and translations.
Most 3DGS-based localization methods follow the classical feature-based visual localization framework. In particular, SplatLoc~\cite{zhai2024splatloc} uses minimal parameters to achieve localization with high-quality rendering.
GSLoc~\cite{liu2024gsloc} establishes 2D-3D correspondences via rendered RGB images and depth maps, enabling localization without training feature descriptors. 
GSplatLoc~\cite{sidorov2024gsplatloc} aligns rendered images with query images by extracting features via XFeat~\cite{potje2024xfeat} for 2D-3D matching during optimization iterations. 
However, like traditional feature-based methods, these methods still suffer from performance degradation in scenes with insufficient texture and structure information.

Our goal is to design a localization method capable of regressing camera poses from arbitrary query images without prior pose. We introduce a multi-level framework that progressively estimates and refines the pose of query image from the global 3DGS map.
Considering the semantic consistency between 2D query image and 3DGS semantic representation, we propose a semantic-based 2D image-to-3DGS matching method. By matching the query image with pre-built 3DGS map, our method provides coarse initial pose estimations. Subsequently, we refine the initial pose via iterative rendering optimization, leveraging the novel view synthesis capability of 3DGS representation.

\begin{figure*}[t] 
    \centering
    \includegraphics[width=\linewidth]{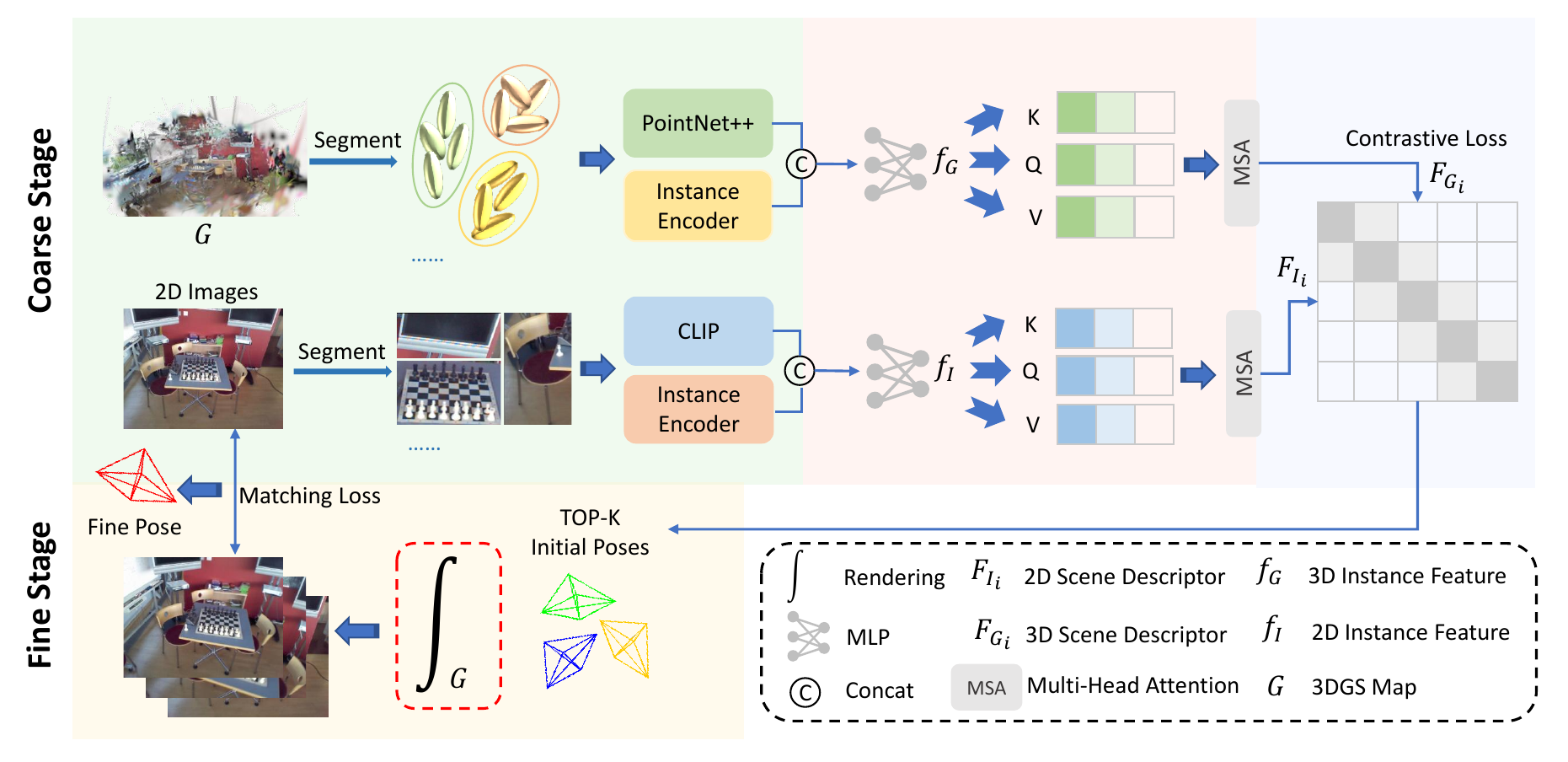} 
    \caption{An overview of SGLoc. Our method takes a query image and 3DGS global map as input. We perform semantic segmentation on both query image and 3DGS representation. 2D and 3D instances are fed into CLIP model~\cite{hafner2021clip} and PointNet++~\cite{qi2017pointnet++} to obtain semantic features respectively. Instance encoders are utilized to encode 2D and 3D instances' color, size, and position information. All features are aggregated as scene semantic descriptors through multi-head attention~\cite{vaswani2017attention} with FFN layer. The semantic-based global retrieval model is guided by contrastive loss to align the 2D and 3D scene semantic descriptors. The top-$k$ submaps are selected by cosine similarity, and corresponding poses are selected as coarse initial poses for pose refinement. Pose is refined through iterative optimization of matching loss between the rendered image and query image.}
    \label{fig:framework}
\end{figure*}

\section{Method}
The overview of our method is shown in Fig.~\ref{fig:framework}. 
We adopt a semantic 3DGS representation~\cite{ye2024gaussian} to obtain 3DGS global map $G$. 
As the query image typically corresponds to a local 3D region rather than the entire scene, we divide the 3DGS map into submaps $G = \{G_{i}: i \in 1,\dots,N\}$. Given a query image $I_q$, we first perform semantic segmentation and extract semantic descriptors from both the image and the 3DGS submaps. 
We define the ground truth camera pose of $I_q$ as $P = [T \mid R]$, where $T \in \mathbb{R}^3$ is the translation vector and $R \in \mathrm{SO}(3)$ is the rotation matrix. 
Then, to provide a reliable coarse initial pose $P^\ast = [T^\ast \mid R^\ast]$ for pose refinement, we align the 3DGS submaps and the query image at the scene level by matching the semantic descriptors $F_I$ of 2D query image and $F_G$ of 3DGS representation. Finally, the coarse pose is further refined by comparing the query image and rendered image $I_r$ from 3DGS representation, resulting in the final estimated pose $\hat{P} = [\hat{T} \mid \hat{R}]$.
Sec.~\ref{sec:Multi-Level} describes our multi-level localization framework.
Sec.~\ref{sec:Global Scene Retrieval} presents our semantic-based global place retrieval. Sec.~\ref{sec:Pose Refinement} introduces details of rendering-based pose refinement.


\subsection{Multi-Level Localization Pipline}
\label{sec:Multi-Level}

We obtain the pose of query image from 3DGS global map in a coarse-to-fine manner.
In the coarse stage, we perform 2D-3D global place retrieval by aligning 2D and 3D scene semantic descriptors into a shared feature space, enabling direct similarity measurement. Through matching 2D and 3D scene semantic descriptors, we retrieve the top-$k$ most similar 3D descriptors corresponding to the query image, which provides $k$ initial poses for downstream optimization.
In the fine stage, we perform rendering-based pose estimation to refine the coarse initial pose.

\noindent\textbf{Coarse Stage.}\hspace{5pt} 
Following the retrieval-based localization strategies introduced in UniLoc~\cite{xia2024uniloc}, we adapt it to a semantic-guided retrieval framework between 2D images and 3DGS representation.
To establish instance-level correspondences, we first perform semantic segmentation on both 2D query image and 3DGS representation. Each 3D submap and 2D image contains multiple object instances $G_i = \{g_j^i:j \in 1,\dots,n\}$, $I = \{p_j:j \in 1,\dots,m\}$. 
The correspondence problem between query image and 3DGS representation is formulated as a retrieval task.
Considering the semantic relationship between 2D query image and 3D scene representation, we extract scene semantic descriptors from the query image and the 3DGS submaps. Then, we map the semantic features into a shared feature space through contrastive learning. 
Moreover, by calculating the similarity scores between scene semantic descriptors of query image and the 3DGS submaps, we identify the top-$k$ submaps that exhibit the highest similarity scores.
The poses corresponding to the top-$k$ candidate submaps are selected as coarse initial poses for subsequent pose refinement. Besides, we filter out obvious retrieval errors before pose refinement by calculating the similarity between the query image and rendered images that are generated from the initial coarse poses.

\noindent\textbf{Fine Stage.}\hspace{5pt} 
Benefitting from the high-quality rendering capability of 3DGS representation~\cite{kerbl20233d}, we leverage the initial coarse poses provided in the first stage to optimize the differences between the query image and the rendered image, obtaining precise pose estimation. 

\subsection{Semantic-based Global Place Retrieval}
\label{sec:Global Scene Retrieval}
\noindent\textbf{Feature Extraction.}\hspace{5pt}
For RGB images, we utilize SAM~\cite{kirillov2023segment} to segment them into instance-level masks. For each segmented instance, we crop the corresponding RGB region based on its mask to obtain an instance-level RGB image. Then, cropped instance images are fed into the CLIP model to extract semantic features $f_{\text{CLIP}} \in \mathbb{R}^{B\times N\times d_c}$, where $B$ denotes the batch size, $N$ is the number of instances, and $d_c$ is the embedding dimension of CLIP model.
$f_{\text{CLIP}}$ is projected into a unified latent space via 3-layer MLP.  
We utilize instance encoder to extract additional instance features. Specifically, for every instance, we encode the average color $\in \mathbb{R}^3$, normalized instance size $\in \mathbb{R}$, and relative position of each instance in UV coordinates $\in \mathbb{R}^2$ through different MLP $\mathcal{F}^{I}_{c}$, $\mathcal{F}^{I}_{s}$, $\mathcal{F}^{I}_{p}$. Then, we concatenate all features and pass through another three-layer MLP to obtain the feature descriptor $f_{I} \in \mathbb{R}^{B\times N\times d}$ for each 2D instance.

For 3DGS submaps, we use 3DGS representation proposed by Gaussian Grouping~\cite{ye2024gaussian} to generate our 3DGS global map. 
~\cite{ye2024gaussian} incorporates new identity encoding parameters to each Gaussian primitive, enabling semantic Gaussian representation.
To extract instance-level 3D features, we employ a pre-trained PointNet++~\cite{qi2017pointnet++} to process the point cloud of each object instance and obtain a semantic embedding $f_{\text{PN}} \in \mathbb{R}^{B\times N\times d}$. The semantic feature is projected into a unified latent space via a learnable 3-layer MLP:
\[
f_{\text{geo}} = \mathcal{F}^{G}_{\text{PN}}(f_{\text{PN}}) \in \mathbb{R}^{d}
\]
Since each 3D Gaussian primitive in the 3DGS model contains both coordinate and color information, we use different MLP $\mathcal{F}^{G}_{c}$, $\mathcal{F}^{G}_{s}$, $\mathcal{F}^{G}_{p}$ to encode the average color $\in \mathbb{R}^3$, the number of 3D Gaussian primitives $\in \mathbb{R}$, and the relative position of each instance projected into the camera coordinate $\in \mathbb{R}^3$. 

All features are integrated through concatenation followed by a three-layer MLP to obtain the feature descriptor $f_{G} \in \mathbb{R}^{B\times N\times d}$.


\noindent\textbf{Feature Aggregation.}\hspace{5pt}  
To establish correspondence between 2D images and 3DGS map, we aggregate instance-level features $f_{I}$, $f_{G}$ into scene semantic descriptors $F_{I}$, $F_{G}$ and then align the scene semantic descriptors from 2D images and 3DGS submaps.
Specifically, to interact with different instance features effectively and assign attention weights to them adaptively, we employ a Multi-Head Self-Attention mechanism~\cite{vaswani2017attention} ($\mathit{Attr}$) with a feed-forward neural network (FFN) for feature aggregation. $\mathit{Attr}$ and the FFN layer take both query (Q), key (K), and value (V) as input. Taking image features as an example, query (Q), key (K) and value (V) are all derived from instance features $f_{I}$. 
\begin{equation}
\begin{aligned}
    \hat{f_{I}} = Q + \mathit{Attr}(Q,K,V), \\
    \hat{F_{I}} = \hat{f_{I}} + FFN(\hat{f_{I}}), \\
    W = \mathit{softmax}(\mathcal{F}(\hat{F_{I}})).
\end{aligned}
\end{equation}

Subsequently, taking $\hat{f_{I}}$ as input, we generate attention weights $W\in\mathbb{R}^{B\times N}$ through a three-layer MLP followed by softmax layer. These attention weights are utilized to aggregate instance descriptors into a scene semantic descriptor: 
\begin{equation}
    F_{I} = \sum_{i=1}^{N}\hat{F_{I_i}}\times{W_i}.
\end{equation}
Here, $W_i$ and $\hat{F_{I_i}}$ denote the attention weight and instance feature corresponding to the $i$-th instance, respectively.

Then, we use cosine similarity to match the 2D and 3D scene semantic descriptors. 
We select the top-$k$ submaps as the result of place retrieval. And the poses corresponding to the top-$k$ submaps are selected as coarse initial poses for pose refinement. 

Since rendered images generated by poses with significant translation and rotation errors diverge substantially from the query image, we employ the Peak Signal-to-Noise Ratio (PSNR)~\cite{johnson2006signal} as the similarity metric to filter out mismatches. PSNR is denoted by the following formula:
\begin{equation}
\begin{aligned}
    \text{PSNR} = 10 \cdot \log_{10} \left( \frac{\text{MAX}_I^2}{\text{MSE}} \right), \\
    \text{MSE} = \frac{1}{hw} \sum_{i=0}^{h-1} \sum_{j=0}^{w-1} \left[ I(i,j) - I_r(i,j) \right]^2,
\end{aligned}
\end{equation}
where $h$ and $w$ represent the height and weight of the image, $I_r$ and $I$ represent the rendered image and query image. $\text{MAX}_I^2$ is the maximum possible pixel value of the image. If PSNR values are below a predefined threshold $\epsilon=55$, we will discard the corresponding initial pose. The filtered coarse initial pose is denoted as ${P^{\ast}_i}=[{T^{\ast}_i}|{R^{\ast}_i}]$

\noindent\textbf{Loss Functions.}\hspace{5pt}
We utilize the contrastive learning loss~\cite{radford2021learning} to align scene semantic descriptors from 3D representation and 2D images. For the $i$-th image and 3D submap pair $(I_i,G_i)$, the contrastive loss function can be calculated using the following formula:
\begin{equation}
\begin{aligned}
    l(I_i,G_i) = f(I_i,G_i)+f(G_i,I_i), \\
        f(I_i,G_i) = -\log \frac{\exp(F_{I_i} \cdot F_{G_i} / \tau)}{\sum_{j \in N} \exp(F_{I_i} \cdot F_{G_j} / \tau)},
\end{aligned}
\end{equation}
where $F_{I_{i}}$ and $F_{G_{i}}$ represent the image and 3D scene semantic descriptors respectively. $\tau$ is the temperature parameter. $N$ is the number of 3DGS submaps in the scene.

The batch loss is derived by averaging the contrastive loss terms.

\subsection{Rendering-based Pose Refinement}

\label{sec:Pose Refinement}
Given a coarse initial pose $P^\ast_{i} = [T^\ast_{i} \mid R^\ast_{i}]$, we adopt a training-free rendering-based method following~\cite{sun2023icomma} to refine pose.
At each optimization step, the image is rendered from the current camera pose. 
Subsequently, the errors between the rendered and query images are calculated, and the camera pose is iteratively refined through gradient-based optimization to minimize this error. 
The problem is formulated as follows:
\begin{equation}
\hat{{P}} = \arg\min \mathcal{L}({I_q}, {I_r|p})
\end{equation}
where $I_r$ is the render image generated by the initial pose ${P^\ast}$, $\hat{{P}}$ presents the predicted pose. We optimize the camera poses by gradient descent. $\mathcal{L}$ is the loss function defined as~\cite{sun2023icomma}, including pixel-level loss $\mathcal{L}_{\text{pixel}}$ and matching loss $\mathcal{L}_{\text{match}}$:
\begin{equation}
\mathcal{L} = \lambda\mathcal{L}_{\text{match}} + (1-\lambda)\mathcal{L}_{\text{pixel}} 
\end{equation}
Where $\lambda$ is the balancing coefficient.
\begin{equation}
\mathcal{L}_{\text{pixel}} = \| I_q - I_r \|_2^2
\end{equation}
\begin{equation}
\mathcal{L}_{\text{match}} = \sum_{k} \| x_k^q - x_k^r \|_2^2
\end{equation}
where \(x_k^q\), \(x_k^r\) are matched keypoints identified by~\cite{sun2021loftr} in the query and rendered images.
The total loss is defined as:

From the top-$k$ initial poses selected by the first stage, the pose associated with the rendered image with the highest similarity to the query image is selected as the final pose $\hat{P} =[{\hat{T}}|{\hat{R}}]$.

\begin{table}[t]
\caption{Accuracy comparison on 12scenes dataset for median translation and rotation errors (cm/\degree) metrics.}
  \centering
    \label{12scene}
    \footnotesize
    \begin{tabular}{c|c c c | c}
    \toprule
    \multirow{2}{*}{Method} & \multicolumn{2}{c}{Apartment 2} & Office 1 & Avg.  $\downarrow$\\
    & Bed & Kitchen & Lounge & [cm/\degree] \\
    \midrule
    SCRNet~\cite{li2020hierarchical} & $3.3/1.5$ & $2.1/1.0$  & $2.7/0.9$ &  $2.7/1.1$ \\
    SCRNet-ID~\cite{ng2021reassessing} &  $2.0/0.8$ & $1.8/0.9$ & $3.4/1.1$ & $2.4/0.9$ \\
    NeRF-SCR~\cite{chen2024leveraging} & $1.6/0.7$  & $1.2/0.5$ & $1.8/0.6$ & $1.5/0.6$  \\
    PNeRFLoc~\cite{zhao2024pnerfloc} & $1.2/0.5$   & $0.8/0.4$  & $2.3/0.8$  & $1.5/0.6$     \\
    SpaltLoc~\cite{zhai2024splatloc} & $1.2/0.5$   & $1.0/0.5$   & $1.6/0.5$   & $1.2/0.5$    \\
    SGLoc (Ours) & \bm{$0.5/0.4$} & \bm{$0.1/0.1$} & \bm{$0.3/0.1$} & \bm{$0.3/0.2$}\\
    \bottomrule
    \end{tabular}
\end{table}

\section{Experiments}
\subsection{Evaluation Setup}
\noindent\textbf{Datasets.}\hspace{5pt}
We evaluate the performance of our SGLoc on two public visual localization datasets, including 4 scenes on 7Scenes dataset~\cite{glocker2013real,shotton2013scene} and 3 scenes on 12Scenes dataset~\cite{radford2021learning}. These datasets contain RGB-D image sequences of various indoor scenes for the evaluation of visual localization performance.

\noindent\textbf{Baselines and Metrics.}\hspace{5pt}
We use \textit{median translation error (cm)} and \textit{rotation error (\degree)} to evaluate the performance of our method. \textit{Avg.} represents the average error.  We compare the metrics with recent traditional localization~\cite{li2020hierarchical,ng2021reassessing,kendall2015posenet,shavit2021learning,chen2022dfnet,chen2024map,brachmann2021visual,brachmann2023accelerated,wang2024glace},
NeRF-based localization~\cite{zhao2024pnerfloc,chen2024leveraging,germain2022feature,moreau2023crossfire,chen2023neural,zhou2024nerfect} and 3DGS-based localization~\cite{zhai2024splatloc,liu2024gsloc} methods.
\noindent\textbf{Implementation Details.}\hspace{5pt}
We use Gaussian Grouping~\cite{ye2024gaussian} to obtain 3DGS global map. 
Each submap is constructed as a cubic region centered around poses sampled from the training trajectories. Specifically, we sample a set of camera poses with a fixed spatial interval depending on the size and complexity of the scene. Overlaps between submaps naturally exist due to the fixed sampling interval.
We train our semantic-based place retrieval using the Adam optimizer. In the coarse stage, we initialize the learning rate (LR) at 1e-3 and train 24 epochs with a batch size of 32. We utilize three-layer MLP and 4-head 2-layer Multi-Head Self-Attention. Besides, $k = 5$ and temperature parameter $\tau = 0.1$. We follow the default settings of all baseline methods to obtain the estimated pose for each query image.

\begin{figure*}[t] 
    \centering
    \includegraphics[width=\linewidth]{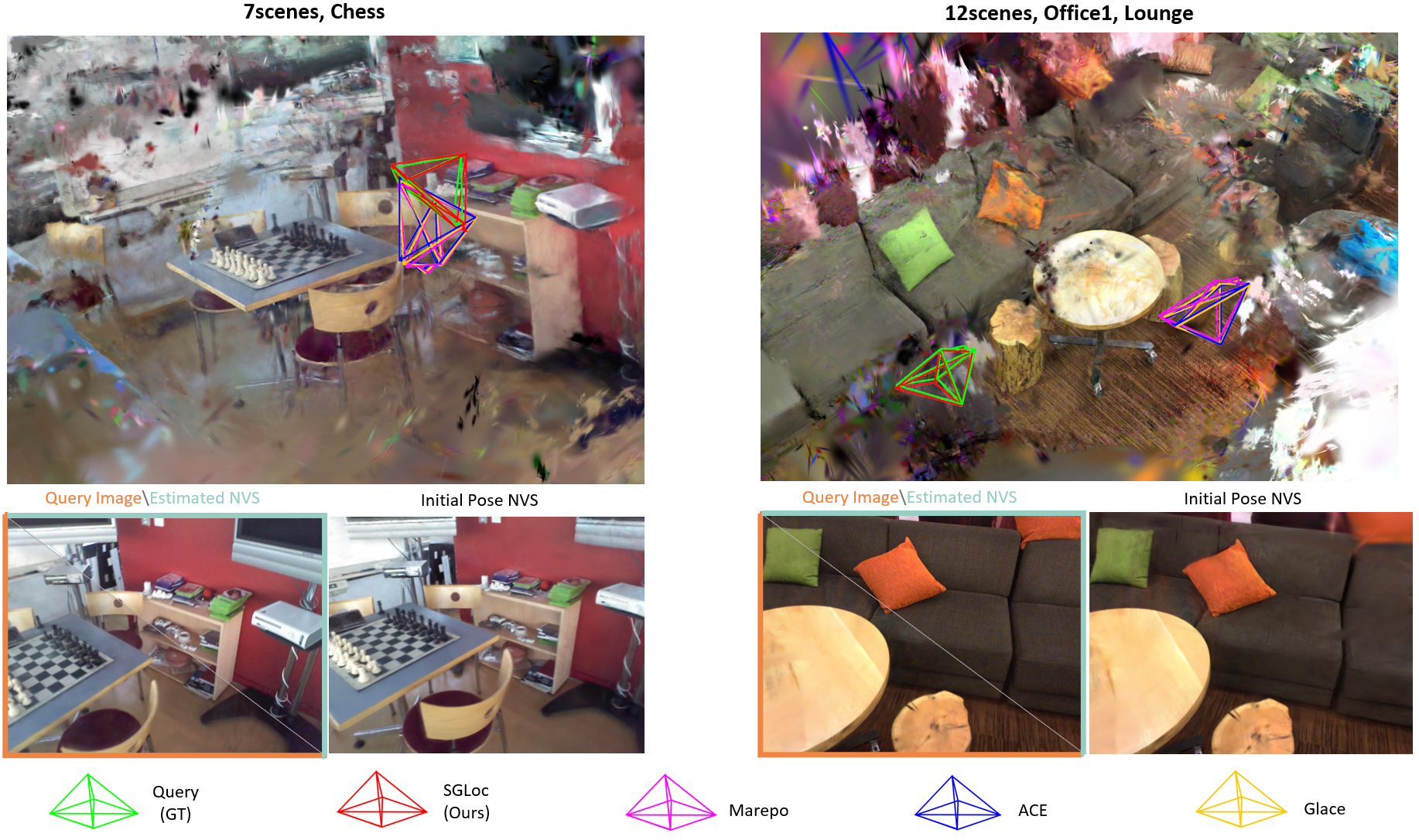} 
    \caption{Qualitative comparison of localization accuracy on the 7Scenes/chess and 12Scenes/lounge scenes. Camera poses with distinct colors represent visualization of initial coarse poses estimated by ACE~\cite{brachmann2023accelerated}, Glace~\cite{wang2024glace}, Marepo~\cite{cheng2022masked}, and our method. (\textcolor[rgb]{1,0.5,0}{Query Image}) Query RGB image;
    (\textcolor[rgb]{0.5,0.8,0.7}{Estimated NVS}) Rendered image using our final estimated pose; (Initial Pose NVS) Rendered image using the initial coarse pose estimated by our method.}
    \label{fig:vis}
\end{figure*}

\begin{table}[t]
\caption{Accuracy comparison on 7scenes dataset for median translation and rotation errors (cm/\degree) metrics.}
\label{7scene}
\centering
\resizebox{\linewidth}{!}{
\begin{tabular}{c|c c c c| c}
\toprule
Method & Chess & Heads & Office & Redkitchen & Avg.$\downarrow$ [cm/\degree]\\
\midrule
PoseNet~\cite{kendall2015posenet} & $10/4.02$ & $18/13.0$ & $17/5.97$ & $22/5.91$ & $16.75/7.23$   \\
MS-Transformer~\cite{shavit2021learning} & $11/6.38$ & $13/13.0$ & $18/8.14$ & $16/8.92$ & $14.5/9.11$   \\
DFNet~\cite{chen2022dfnet} &  $3/1.12$ & $4/2.29$ & $6/1.54$ & $7/1.74$ & $5/1.67$  \\
Marepo~\cite{cheng2022masked} & $1.9/0.83$ & $2.1/1.24$ & $2.9/0.93$ & $2.9/0.98$ & $2.45/1.0$       \\
DSAC*~\cite{brachmann2017dsac} & $0.5/0.17$ & $0.5/0.34$ & $1.2/0.34$ & $\bm{0.7}/0.21$ & $0.73/0.27$          \\
ACE~\cite{brachmann2023accelerated} & $0.5/0.18$ & $0.5/0.33$ & $1/0.29$ & $0.8/0.20$ & $0.7/0.5$        \\
GLACE~\cite{wang2024glace} & $0.6/0.18$ & $0.6/0.34$ & $1.1/0.29$ & $0.8/0.20$ & $0.78/0.25$        \\
FQN-MN~\cite{germain2022feature} & $4.1/1.31$ & $9.2/2.45$ & $3.6/2.36$ & $16.1/4.42$ & $8.25/2.64$         \\
CrossFire~\cite{moreau2023crossfire} & $1/0.4$ & $3/2.3$ & $5/1.6$ & $2/0.8$ & $2.75/1.28$     \\
DFNet + NeFeS\textsubscript{50}~\cite{chen2023neural} & $2/0.57$ & $2/1.28$ & $2/0.56$ & $2/0.57$ & $2.1/0.75$         \\
HR-APR~\cite{liu2024hr} & $2/0.5$5 & $2/1.45$ & $2/0.64$ & $2/0.67$ & $2/0.82$    \\
NeRFMatch~\cite{zhou2024nerfect} & $0.9/0.3$ & $1.6/1.0$ & $3.3/0.7$ & $1.3/0.3$ & $1.78/0.58$      \\
DFNet + GSLoc~\cite{liu2024gsloc} & $1.3/0.35$ & $1.1/0.71$ & $2.2/0.5$ & $2.2/0.47$ & $1.7/0.51$    \\
Marepo + GSLoc~\cite{liu2024gsloc} &  $1.3/0.4$ & $1.4/0.68$ & $2.2/0.5$ & $2.2/0.48$ & $1.78/0.52$       \\
ACE + GSLoc~\cite{liu2024gsloc} & $0.5/0.15$ & $0.5/0.28$ & $1/0.25$ & $0.8/\bm{0.17}$ & $0.7/0.21$     \\
SGLoc (Ours) & \bm{$0.14/0.05$} & \bm{$0.14/0.06$} & \bm{$0.43/0.22$} & $1.3/0.26$ & \bm{$0.5/0.15$}    \\
\bottomrule
\end{tabular}}
\end{table}

\begin{table}[t]
\caption{Ablation study of using different initial pose estimators on 12scenes dataset.}
\label{ab12}
\resizebox{\linewidth}{!}{
\begin{tabular}{c|c c c|c}
\toprule
\multirow{2}{*}{Method} & \multicolumn{2}{c}{Apartment 2} & Office 1 & Avg. $\downarrow$\\
  & Bed & Kitchen &  Lounge  & [cm/\degree]\\
\midrule
ACE\cite{brachmann2023accelerated}+ SGLoc$_2$ & $610.44/73.77$ & $152.27/150.40$ & $147.48/97.62$  & $303.40/107.26$   \\
GLACE\cite{wang2024glace}+ SGLoc$_2$ & $500.557/76.49$ & $139.38/85.55$ & $118.22/113.83$  & $252.72/91.96$   \\
Marepo\cite{cheng2022masked}+ SGLoc$_2$ &  $518.10/79.70$ & $97.70/47.24$ & $257.61/174.91$  & $299.14/100.62$  \\
SGLoc (Ours) & $\bm{0.48/0.39}$ & $\bm{0.11/0.05}$ & $\bm{0.28/0.08}$  & $\bm{0.29/0.17}$    \\
\bottomrule
\end{tabular}}
\end{table}

\begin{table}[t]
\caption{Ablation study of using different initial pose estimators on 7scenes dataset.}
\label{ab7}
\resizebox{\linewidth}{!}{
\begin{tabular}{c|c c c c| c}
\toprule
Method & Chess & Heads & Office  & Redkitchen & Avg.$\downarrow$ [cm/\degree] \\
\midrule
ACE\cite{brachmann2023accelerated}+ SGLoc$_2$ & $186.29/68.13$ & $67.41/70.80$ & $225.67/79.73$ & $417.51/49.54$ & $224.22/67.05$   \\
GLACE\cite{wang2024glace}+ SGLoc$_2$ & $266.19/174.91$ & $106.29/85.55$ & $226.56/80.99$ & $339.40/50.73$ & $234.71/98.05$   \\
Marepo\cite{cheng2022masked}+ SGLoc$_2$ &  $147.04/171.36$ & $66.97/77.26$ & $139.58/108.26$ & $335.70/46.94$ & $172.32/100.96$  \\
SGLoc (Ours) & $\bm{0.14/0.05}$ & $\bm{0.14/0.06}$ & $\bm{0.43/0.22}$ & $\bm{1.3/0.26}$ & $\bm{0.5/0.15}$    \\
\bottomrule
\end{tabular}}
\end{table}

\subsection{Experimental Results}
\noindent\textbf{Localization Results.}\hspace{5pt}
As shown in Tab.~\ref{12scene}, our method outperforms other baseline methods in 12Scenes dataset~\cite{radford2021learning}, as well as achieves up to 87.5\% increase in translation accuracy and 80\% increase in rotation accuracy. Tab.~\ref{7scene} demonstrates that our method achieves the highest average accuracy in 7scenes dataset~\cite{glocker2013real, shotton2013scene}, with the lowest average translation ($0.15$cm) and rotation ($0.05\degree$) errors. Moreover, our method achieves 29\% relative increase in average median translation and rotation errors. 
Such improvement is attributed to our semantic-based global retrieval method, which provides precise initial poses for pose regression.
By leveraging semantic consistency to establish correspondence between query image and 3DGS map, our method achieves superior performance over other methods that are based on the traditional feature extraction.


\noindent\textbf{Visualization Results.}\hspace{5pt}
To further demonstrate the effectiveness of our approach, we visualize the localization comparison results of 2 scenes in Fig.~\ref{fig:vis}.
The visualization of each scene contains three components: (1) a subfigure of the query image and the rendered image using the estimated pose (bottom left), (2) visualization of initial coarse poses estimated by ACE~\cite{brachmann2023accelerated}, Glace~\cite{wang2024glace}, Marepo~\cite{cheng2022masked}, and our method (top panel, with distinct colors), (3) a rendered image 
generated from the initial coarse pose estimated by our method (bottom right). In the subfigure (bottom left), a diagonal line divides into 2 parts: the bottom-left quadrant displays the query image, while the top-right quadrant shows the rendered image with our estimated pose.
As is shown in Fig.~\ref{fig:vis}, initial poses provided by our method are the closest to the ground truth, which fully demonstrates the accuracy of our designed coarse pose estimator. The initial poses estimated by other methods lead to large errors, especially in 12scenes/lounge scene. 
This improvement is attributed to our semantic-based global place retrieval strategy that leverages semantic consistency between 2D query image and 3DGS global map to directly obtain initial pose estimation.


\subsection{Ablation Studies}
In this section, we validate the effectiveness of our semantic-based place retrieval module and demonstrate that our rendering-based optimization can effectively achieve pose refinement.

\noindent\textbf{Effects of semantic-based global retrieval.}\hspace{5pt}
To evaluate the effectiveness of our semantic-based global retrieval algorithm, we employ initial poses predicted by three state-of-the-art pose estimators (ACE~\cite{brachmann2023accelerated}, Glace~\cite{wang2024glace}, Marepo~\cite{chen2024map}) as input for pose refinement. SGLoc$_2$ denotes our rendering-based pose refinement module. The localization performance is evaluated by median rotation error (\degree) and translation error ($cm$) metrics.
As shown in Tab.~\ref{ab12} and Tab.~\ref{ab7}, we present localization results with different initialization strategies. Experimental results demonstrate that these coarse pose estimators followed by the same pose refinement module generally fail to accomplish localization tasks on two datasets. However, our global retrieval algorithm achieves superior performance. It also indicates that our semantic-based global location retrieval module has the most powerful matching capability and robustness in various scenes, which are attributed to full extraction and integration of global semantic features.

\noindent\textbf{Effects of rendering-based pose refinement.}\hspace{5pt}
As shown in Tab.~\ref{fine12} and Tab.~\ref{fine7}, rendering-based optimization can effectively reduce the translation error and rotation error by at least 5 times and can even reach the error level of 0.1 $cm$ and 0.01\degree. It also demonstrates that, given a better coarse initial pose, rendering-based optimization can achieve accurate localization results without the need for designing a more complex pose refinement strategy.

\begin{table}[t]
\caption{Ablation study of our SGLoc on the 12scenes dataset.'w/o SGLoc$_2$' indicates without our pose refinement module.}
\label{fine12}
\centering
\footnotesize
\begin{tabular}{c|c c c | c}
\toprule
\multirow{2}{*}{Method} & \multicolumn{2}{c}{Apartment 2} & Office 1 &  Avg.$\downarrow$ \\
& Bed & Kitchen &  Lounge & [$cm$/\degree]\\
\midrule
w/o SGLoc$_2$  & $4.26/1.82$ & $1.58/5.35$  & $2.96/5.21$   & $4.24/4.52$ \\
SGLoc (Ours) &  $\bm{0.48/0.39}$ & $\bm{0.11/0.05}$ & $\bm{0.28/0.08}$  & $\bm{0.29/0.17}$    \\
\bottomrule
\end{tabular}
\end{table}

\begin{table}[t]
\caption{Ablation study of our SGLoc on the 7scenes dataset. 'w/o SGLoc$_2$' indicates without our pose refinement module.}
\label{fine7}
\centering
\resizebox{\linewidth}{!}{
\begin{tabular}{c|c c c c| c}
\toprule
Method & Chess &  Heads & Office & Redkitchen  & Avg.$\downarrow$ [$cm$/\degree]\\
\midrule
w/o SGLoc$_2$ & $2.64/0.44$ & $5.4/0.52$ & $1.57/3.12$ & $6.26/5.42$ & $3.97/2.38$ \\
SGLoc (Ours) & $\bm{0.14/0.05}$ & $\bm{0.14/0.06}$ & $\bm{0.43/0.22}$ & $\bm{1.3/0.26}$ & $\bm{0.5/0.15}$    \\
\bottomrule
\end{tabular}}
\end{table}

\section{CONCLUSIONS}
We propose SGLoc, a novel localization framework that estimates 6DoF pose from 3D Gaussian Splatting (3DGS) representation through semantic information.
By designing a multi-level localization strategy guided by semantic consistency, our method achieves competitive global localization effects without prior pose information. We introduce a semantic-based global retrieval algorithm that aligns the image with the local region of the global 3DGS map to obtain a coarse pose estimation. Subsequently, we perform rendering-based pose refinement through iterative optimization of the differences between the query image and the rendered image from 3DGS. Experiments demonstrate that our SGLoc achieves superior performance over baselines on 12scenes and 7scenes datasets, showing excellent capabilities in global localization without initial pose prior.


\bibliographystyle{IEEEtran}  
\bibliography{IEEEabrv,root} 

\end{document}